\documentclass[11pt,a4paper]{article}
\usepackage[hyperref]{acl2020}
\usepackage{times}
\usepackage{latexsym}
\usepackage{booktabs}       
\usepackage{dblfloatfix} 
\usepackage{graphicx}

\usepackage{microtype}

\aclfinalcopy 

\newcommand\BibTeX{B\textsc{ib}\TeX}

\title{Audio-Visual Understanding of Passenger Intents for~In-Cabin~Conversational~Agents}

\author{Eda Okur \qquad Shachi H Kumar \qquad Saurav Sahay \qquad Lama Nachman \\
  Intel Labs, Anticipatory Computing Lab, USA \\
  \texttt{\{eda.okur, shachi.h.kumar, saurav.sahay, lama.nachman\}@intel.com} \\}

\date{}

\begin{document}
\maketitle
\begin{abstract}
Building multimodal dialogue understanding capabilities situated in the in-cabin context is crucial to enhance passenger comfort in autonomous vehicle (AV) interaction systems. To this end, understanding passenger intents from spoken interactions and vehicle vision systems is a crucial component for developing contextual and visually grounded conversational agents for AV. Towards this goal, we explore AMIE (Automated-vehicle Multimodal In-cabin Experience), the in-cabin agent responsible for handling multimodal passenger-vehicle interactions. In this work, we discuss the benefits of a multimodal understanding of in-cabin utterances by incorporating verbal/language input together with the non-verbal/acoustic and visual clues from inside and outside the vehicle. Our experimental results outperformed text-only baselines as we achieved improved performances for intent detection with a multimodal approach.
\end{abstract}

\section{Introduction}

Understanding passenger intents from spoken interactions and visual cues (both from inside and outside the vehicle) is an important building block towards developing contextual and scene-aware dialogue systems for autonomous vehicles. When the passengers give instructions to the in-cabin agent AMIE, the agent should parse commands properly considering three modalities (i.e., verbal/language/text, vocal/audio, visual/video) and trigger the appropriate functionality of the AV system. 

For in-cabin dialogue between car assistants and driver/passengers, recent studies explore creating a public dataset using a WoZ approach \cite{stanford-2017} and improving ASR for passenger speech recognition \cite{asr-fukui-2018}. Another recent work \cite{Zheng-2017} attempts to classify sentences as navigation-related or not using the CU-Move in-vehicle speech corpus \cite{CU-Move-2001}, a relatively old and large corpus focusing on route navigation. 

We collected a multimodal in-cabin dataset with multi-turn dialogues between the passengers and AMIE using a Wizard-of-Oz (WoZ) scheme via realistic scavenger hunt game. In previous work~\cite{AMIE-CICLing-2019}, we experimented with various RNN-based models to detect the utterance-level intents (i.e., \textit{set-destination}, \textit{change-route}, \textit{go-faster}, \textit{go-slower}, \textit{stop}, \textit{park}, \textit{pull-over}, \textit{drop-off}, \textit{open-door}, \textit{other}) along with the intent keywords and relevant slots (i.e., \textit{location}, \textit{position/direction}, \textit{object}, \textit{gesture/gaze}, \textit{time-guidance}, \textit{person}) associated with these intents. 

In this work, we discuss the benefits of a multimodal understanding of in-cabin utterances by incorporating verbal/language input together with the non-verbal/acoustic and visual cues, both from inside and outside the vehicle (e.g., passenger gestures and gaze from the in-cabin video stream, referred objects outside of the vehicle from the road view camera stream). 

\section{Data}

\begin{figure}[t!]
\includegraphics[width=\columnwidth]{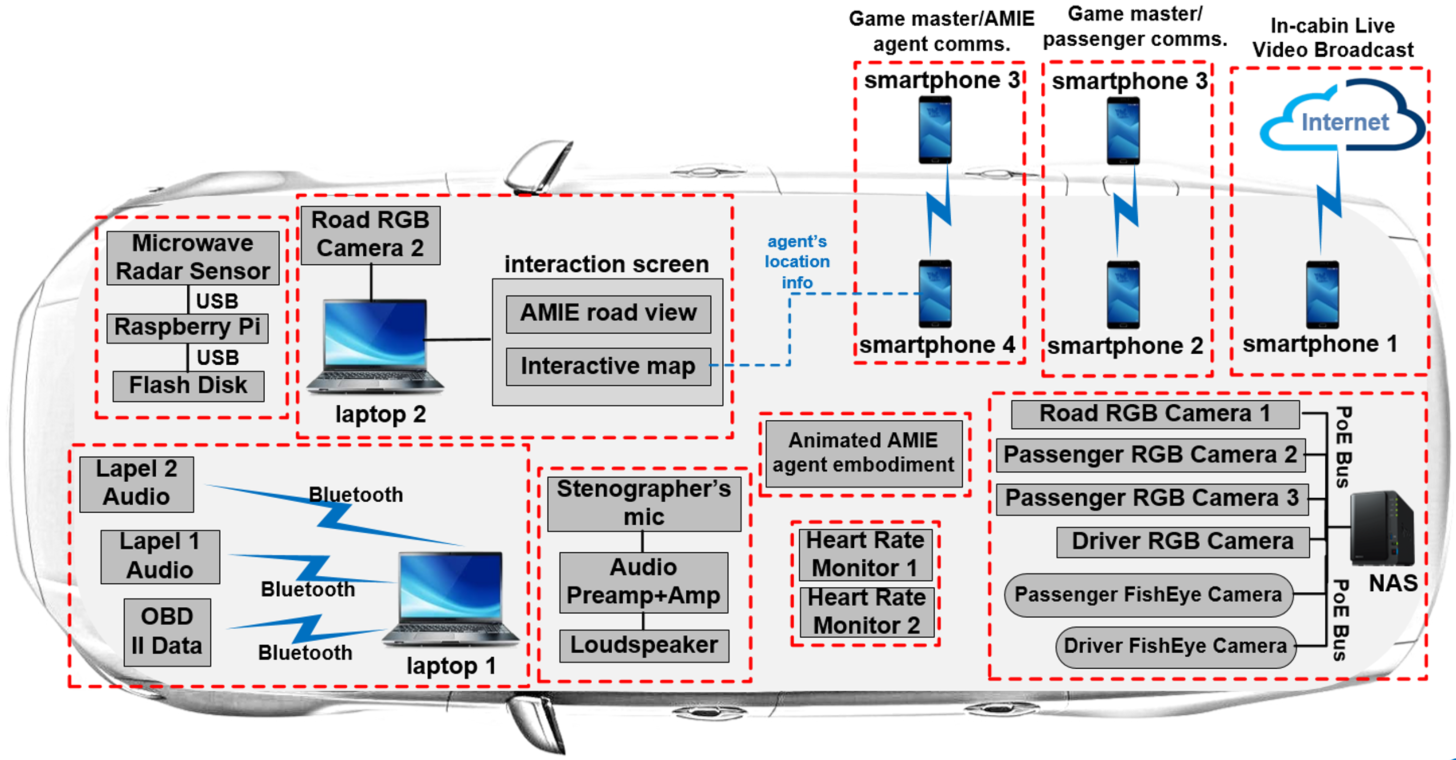}
\caption{AMIE In-cabin Data Collection Setup}
\label{fig:car}
\end{figure}

Our AMIE in-cabin dataset includes 30 hours of multimodal data collected from 30 passengers (15 female, 15 male) in a total of 20 sessions. In 10 sessions, a single passenger was present, whereas the remaining 10 sessions include two passengers interacting with the vehicle. Participants sit in the back of the vehicle, separated from the driver and the human acting as an agent at the front. The vehicle is modified to hide the operator and the WoZ AMIE agent from the passengers, using a variation of the WoZ approach \cite{WoZ-2017}. In each ride/session, which lasted about 1 hour or more, the participants were playing a realistic scavenger hunt game on the streets of Richmond, BC, Canada. Passengers treat the vehicle as AV and communicate with the WoZ AMIE agent mainly via speech commands. Game objectives require passengers to interact naturally with the agent to go to certain destinations, update routes, give specific directions regarding where to pull over or park (sometimes with gestures), find landmarks (refer to outside objects), stop the vehicle, change speed, get in and out of the vehicle, etc. Further details of the data collection protocol and dataset statistics can be found in~\cite{HRI-2018, AMIE-CICLing-2019}. See Fig.~\ref{fig:car} for the vehicle instrumentation to enable multimodal data collection setup.

\subsection{Dataset Statistics}

Multimodal AMIE dataset consists of in-cabin conversations between the passengers and the AV agent, with 10590 utterances in total. 1331 of these utterances have commands to the WoZ agent; hence, they are associated with passenger intents. Utterance-level intent and word-level slot annotations are obtained on the transcribed utterances by majority voting of 3 annotators. The annotation results for \textit{utterance-level intent} types, \textit{slots} and \textit{intent keywords} can be found in Table~\ref{D1} and Table~\ref{D2}.

\begin{table}[b!]
  \centering
  \resizebox{\columnwidth}{!}{
  \begin{tabular}{*3c}
    \toprule
    \textbf{AMIE Scenario} & \textbf{Intent Type} & \textbf{Utterance Count} \\
    \toprule
    Set/Change & SetDestination & 311 \\
    Destination/Route & SetRoute & 507 \\
    \midrule
     & Park & 151 \\
    Finishing the Trip & PullOver & 34 \\
     & Stop & 27 \\
    \midrule
    Set/Change & GoFaster & 73 \\
    Driving Behavior/Speed & GoSlower & 41 \\
    \midrule
    Others & OpenDoor & 136 \\
    (Door, Music, A/C, etc.) & Other & 51 \\
    \midrule
     & \textit{Total} & \textit{1331} \\
    \bottomrule
  \end{tabular}
  }
  \caption{AMIE In-cabin Dataset Statistics: Intents}
  \label{D1}
\end{table}

\begin{table}[b!]
  \centering
  \resizebox{0.65\columnwidth}{!}{
  \begin{tabular}{*2c}
    \toprule
    \textbf{Slot/Keyword Type} & \textbf{Word Count} \\
    \toprule
    Intent Keyword & 2007 \\
    \midrule
    Location & 1969 \\
    Position/Direction & 1131 \\
    Person & 404 \\
    Time Guidance & 246 \\
    Gesture/Gaze & 167 \\
    Object & 110 \\
    \midrule
    None & 6512 \\
    \midrule
    \textit{Total} & \textit{12546} \\
    \bottomrule
  \end{tabular}
  }
  \caption{AMIE In-cabin Dataset Statistics: Slots}
  \label{D2}
\end{table}

\section {Methodology}

We explored leveraging multimodality for the Natural Language Understanding (NLU) module in the Spoken Dialogue System (SDS) pipeline. As our AMIE in-cabin dataset has audio and video recordings, we investigated three modalities for the NLU: text, audio, and visual.

For text (verbal/language) modality, we employed the Hierarchical \& Joint Bi-LSTM model~\cite{bi-lstm-1997,hakkani-2016,zhang-2016,wen-2018}, namely H-Joint-2.

\begin{itemize}
    \item \textbf{Hierarchical \& Joint Model (H-Joint-2)}: This is a 2-level hierarchical joint learning model that detects/extracts \textit{intent keywords} \& \textit{slots} using sequence-to-sequence Bi-LSTMs first (Level-1), then only the words that are predicted as \textit{intent keywords} \& \textit{valid slots} are fed into the Joint-2 model (Level-2), which is another sequence-to-sequence Bi-LSTM network for \textit{utterance-level intent} detection, jointly trained with \textit{slots} \& \textit{intent keywords}.
\end{itemize}

This architecture was chosen based on the best-performing uni-modal results presented in previous work~\cite{AMIE-CICLing-2019} for utterance-level intent recognition and slot filling on our AMIE dataset. These initial uni-modal results were obtained on the transcribed text with pre-trained GloVe word embeddings~\cite{glove-2014}.

In this study, we explore the following multimodal features to better assess passenger intents for conversational agents in self-driving cars: word embeddings for text, speech embeddings and acoustic features for audio, and visual features for the video modality.

\begin{table*}[!t]
  \centering
  \begin{tabular}{lll}
    \toprule
    \textbf{Modalities} & \textbf{Features} & \textbf{F1}(\%) \\
    \midrule
Text & Word2Vec (37.6K vocab) & 85.63 \\
Text & GloVe (400K vocab) & 89.02 \\
Text \& Audio & GloVe \& Acoustic (openSMILE/IS10) & 89.53 \\
Text \& Visual & GloVe \& Video\_cabin (CNN/Inception-ResNet) & 89.40 \\
Text \& Visual & GloVe \& Video\_road (CNN/Inception-ResNet) & 89.37 \\
Text \& Visual & GloVe \& Video\_cabin+road (CNN/Inception-ResNet) & 89.68 \\
\midrule
Audio & Speech2Vec (37.6K vocab) & 84.47 \\
Text \& Audio & Word2Vec+Speech2Vec & 88.08 \\
Text \& Audio & GloVe+Speech2Vec & 90.85 \\
Text \& Audio & GloVe+Word2Vec+Speech2Vec & 91.29 \\
Text \& Audio & GloVe+Word2Vec+Speech2Vec \& Acoustic (IS10) & 91.68 \\
Text \& Audio \& Visual & GloVe+Word2Vec+Speech2Vec \& Video\_cabin (CNN) & 91.50 \\
Text \& Audio \& Visual & GloVe+Word2Vec+Speech2Vec \& Video\_cabin+road (CNN) & 91.55 \\
    \bottomrule
  \end{tabular}
  \caption{F1-scores of Intent Recognition with Multimodal Features}
  \label{audio-video}
\end{table*}

\subsection{Word and Speech Embeddings}
We incorporated pre-trained speech embeddings, called Speech2Vec\footnote{\url{https://github.com/iamyuanchung/speech2vec-pretrained-vectors}}, as additional audio-related features. These Speech2Vec embeddings~\cite{speech2vec-Chung2018} are trained on a corpus of~500 hours of speech from LibriSpeech. Speech2Vec can be considered as a speech version of Word2Vec embeddings~\cite{word2vec-nips-2013}, where the idea is that learning the representations directly from speech can capture the information carried by speech that may not exist in plain text. 

We experimented with concatenating word and speech vectors using GloVe embeddings (6B tokens, 400K vocab, 100-dim), Speech2Vec embeddings (37.6K vocab, 100-dim), and its Word2Vec (37.6K vocab, 100-dim) counterpart, in which the Word2Vec embeddings are trained on the transcript of the same LibriSpeech corpus.

\subsection{Acoustic Features}
Using openSMILE\footnote{\url{https://www.audeering.com/opensmile/}} audio feature extraction toolkit~\cite{openSMILE-2013}, 1582 acoustic features are extracted for each utterance using the segmented audio clips from AMIE dataset. These are the INTERSPEECH 2010 Paralinguistic Challenge (IS10) features~\cite{IS10-schuller2010interspeech} including PCM (pulse-code modulation) loudness, MFCC (Mel-frequency cepstral coefficients), log Mel Freq. Band, LSP (line spectral pairs) Frequency, etc.

\subsection{Visual Features}
Intermediate CNN features\footnote{\url{https://github.com/MKLab-ITI/intermediate-cnn-features}} are extracted from each video clip segmented per utterance from the AMIE dataset. Using the feature extraction process described in~\cite{cnn-feats-kordopatis2017}, one frame per second is sampled for any given input video clip. Then, its visual descriptors are extracted from the activations of the intermediate convolution layers of a pre-trained CNN. 

We used the pre-trained Inception-ResNet-v2 model\footnote{\url{https://github.com/tensorflow/models/tree/master/research/slim}}~\cite{inception-SzegedyIV16} and generated 4096-dim features for each sample (per utterance). We experimented with utilizing two sources of visual information: (i) cabin/passenger view from the back-driver RGB camera recordings, (ii) road/outside view from the dash-cam RGB video streams.

\section{Experimental Results}


Performance results of the utterance-level intent recognition models with various modality and feature concatenations can be found in Table~\ref{audio-video}, using hierarchical joint learning (H-Joint-2). For text and speech embeddings experiments, we observe that using Word2Vec or Speech2Vec representations achieve comparable F1-score performances, which are significantly below the GloVe embeddings performance. That was expected as the pre-trained Speech2Vec vectors have lower vocabulary coverage than the GloVe vectors. On the other hand, we observe that concatenating GloVe + Speech2Vec embeddings, and further GloVe + Word2Vec + Speech2Vec yields higher F1-scores for intent recognition. These results show that the speech embeddings indeed can capture useful semantic information carried by the speech only, which may not exist in plain text.

We also investigate incorporating the audio-visual features on top of text-only and text + speech embedding models. Including openSMILE/IS10 acoustic features from audio as well as intermediate CNN/Inception-ResNet-v2 features from video brings slight improvements to our intent recognition models, achieving 0.92 F1-score. These initial results may require further explorations for specific intents such as \textit{stop} (e.g., audio intensity \& loudness could have helped), or for relevant slots such as passenger \textit{gesture/gaze} (e.g., cabin-view features) and outside \textit{object}s (e.g., road-view features).

\section{Conclusion and Future Work}

In this work, we briefly present our initial explorations towards the multimodal understanding of passenger utterances in autonomous vehicles. We show that our experimental results outperformed the uni-modal text-only baseline results, and with multimodality, we achieved improved performances for passenger intent detection in AV. This ongoing research has the potential impact of exploring real-world challenges with human-vehicle-scene interactions for autonomous driving support via spoken utterances.

There exist various exciting recent work on improved multimodal fusion techniques~\cite{zadeh2018memory, liang-etal-2019-strong, pham2019found, 8269806}. In addition to the simplified feature and modality concatenations, we plan to explore some of these promising tensor-based multimodal fusion networks~\cite{Liu_2018, liang-etal-2019-learning, tsai2019MULT} for more robust intent classification on AMIE dataset as future work. 


\bibliography{anthology,acl2020}
\bibliographystyle{acl_natbib}

\end{document}